\documentclass[peerreview, onecolumn]{IEEEtran}

\usepackage{url} 
\usepackage[utf8]{inputenc} 
\usepackage{booktabs} 
\usepackage{graphicx}
\usepackage{orcidlink}
\usepackage{svg}
\usepackage{float} 
\usepackage{caption}
\usepackage{biblatex}
\usepackage{makecell}
\usepackage{bbding}
\usepackage{pifont}
\usepackage{wasysym}
\usepackage{amssymb}
\usepackage[affil-it]{authblk}
\usepackage[english]{babel}
\usepackage{blindtext}
\usepackage{geometry}
\begin{document}
\newgeometry{top=5cm, bottom=3cm, left=3cm, right=3cm}
\title{\huge GPT-4V as Traffic Assistant: An In-depth Look at Vision Language Model on Complex Traffic Events}
\restoregeometry




\author[1,*]{Xingcheng Zhou\orcidlink{0000-0003-1178-5221},}
\author[1]{Alois C. Knoll\orcidlink{0000-0003-4840-076X}}

\affil[1]{Technical University of Munich, Germany}
\affil[*]{Corresponding author: xingcheng.zhou@tum.de}


\maketitle
\tableofcontents
\listoffigures
\listoftables

\IEEEpeerreviewmaketitle
\begin{abstract}

The recognition and understanding of traffic incidents, particularly traffic accidents, is a topic of paramount importance in the realm of intelligent transportation systems and intelligent vehicles. This area has continually captured the extensive focus of both the academic and industrial sectors. Identifying and comprehending complex traffic events is highly challenging, primarily due to the intricate nature of traffic environments, diverse observational perspectives, and the multifaceted causes of accidents. These factors have persistently impeded the development of effective solutions. The advent of large vision-language models (VLMs) such as GPT-4V, has introduced innovative approaches to addressing this issue.  In this paper, we explore the ability of GPT-4V with a set of representative traffic incident videos and delve into the model's capacity of understanding these complex traffic situations. We observe that GPT-4V demonstrates remarkable cognitive, reasoning, and decision-making ability in certain classic traffic events. Concurrently, we also identify certain limitations of GPT-4V, which constrain its understanding in more intricate scenarios. These limitations merit further exploration and resolution.

\end{abstract}

\section{Introduction}

The existing traffic event recognition approaches \cite{xu2021sutdtrafficqa,chen2023temadapter} do not work well in practice due to the lack of common sense, high-level reasoning ability, and poor zero-shot recognition ability. Large Vision Language Models (VLMs) \cite{liu2023llava,zhu2023minigpt4,geminiteam2023gemini}, represented by GPT-4V \cite{GPT4V}, integrate the vision modality with large language models (LLMs). They inherit the emergent abilities of LLMs \cite{GPT4,touvron2023llama,touvron2023llama2}, including strong reasoning and understanding abilities, the capability to leverage common sense, and zero-shot recognition abilities. It provides the possibility of accomplishing complex multimodal tasks as well as tackling the challenges in traffic event recognition. GPT-4V, recognized as one of the most powerful multi-modal models to date, has garnered widespread attention since its release. Several papers have shown exploratory applications of GPT-4V in different domains \cite{wen2023road,zhou2023visionlminadandits,yang2023dawn}. Therefore we are curious whether it can address these issues and bring new solution to this field. \\


\begin{figure}[H]
\centering
\includegraphics[width=0.9\linewidth]{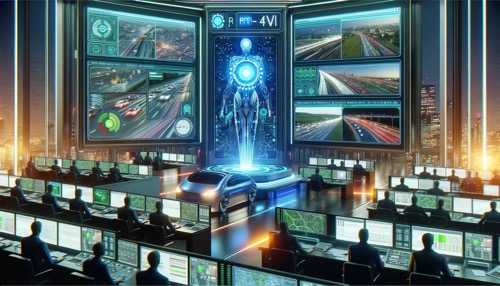}
\caption*{Fig. Image Generated with DALLE-3 by prompting our paper title: \\ "GPT-4V as Traffic Assistant: An In-depth Look at Vision Language Model on Complex Traffic Events"}
\end{figure}

In our study, we qualitatively evaluate how much GPT-4V understands various traffic events. To ensure the diversity and representatives of the evaluation, we carefully selected the key frames from a variety of videos showing classic traffic events. The selections are made with the objective of providing a complete and detailed overview of each incident. It means that chosen frames contain the essential information required for logical analysis and reasoning of the events depicted. The evaluated videos come from a wide range of sources, including the \textbf{Providetia++} infrastructure \cite{creß2022a9dataset,providentia++}, open-source datasets\cite{tad,xu2021sutdtrafficqa}, and various Internet resources. For clarity, the dashed boxes on the images are marked to help the reader comprehend the traffic events more quickly, and are not given as additional hints to GPT-4V. We set the temperature as 0 during evaluation to achieve the most confident and precise results from the model. 







\section{Success Cases}
In this section, we present several exemplary success cases during the evaluations. We classify those traffic incidents that GPT-4V successfully recognizes both the presence of traffic events and the traffic type as success cases, although some of these instances may encompass certain misconceptions in event understanding.

\subsection{Dooring}
\label{Dooring}
\begin{figure}[H]
\centering
\includegraphics[width=\linewidth]{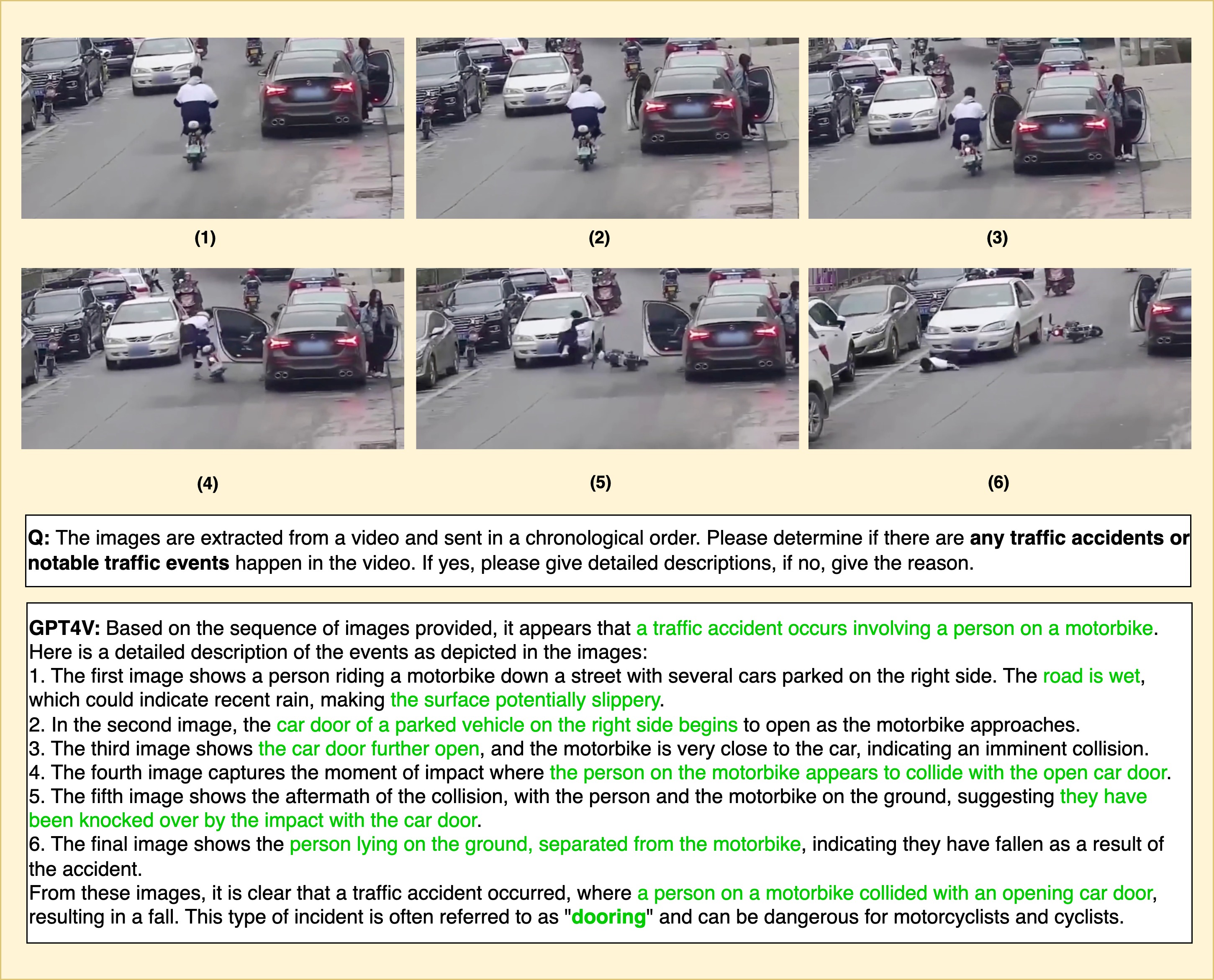}
\caption{Dooring Part 1}
\label{Dooring Part 1}
\end{figure}

Dooring is a common type of traffic accident, usually occurring in areas with dense traffic, such as urban settings. It typically happens when a vehicle is parked roadside, and a passenger or driver inside the vehicle opens the door without noticing an approaching cyclist or motorcyclist. The sudden opening of the door often leaves the approaching individuals without sufficient time to avoid the collision, which can lead to an accident. It is usually very risky for vulnerable road users. The six images in Figure \ref{Dooring Part 1} are extracted from a video resourced from \cite{Dooring}. It illustrates a traffic accident where an electric scooter rider was pushed and fell to the side by a car door suddenly opening. We first evaluate whether GPT-4V can recognize the traffic accident occurring from the keyframes, and then test its performance in generating traffic accident reports, which is a critical task in traffic applications. We also evaluate model's abilities of causal reasoning, attribution, and decision-making in such a complex traffic event.\\

\begin{figure}[H]
\centering
\includegraphics[width=\linewidth]{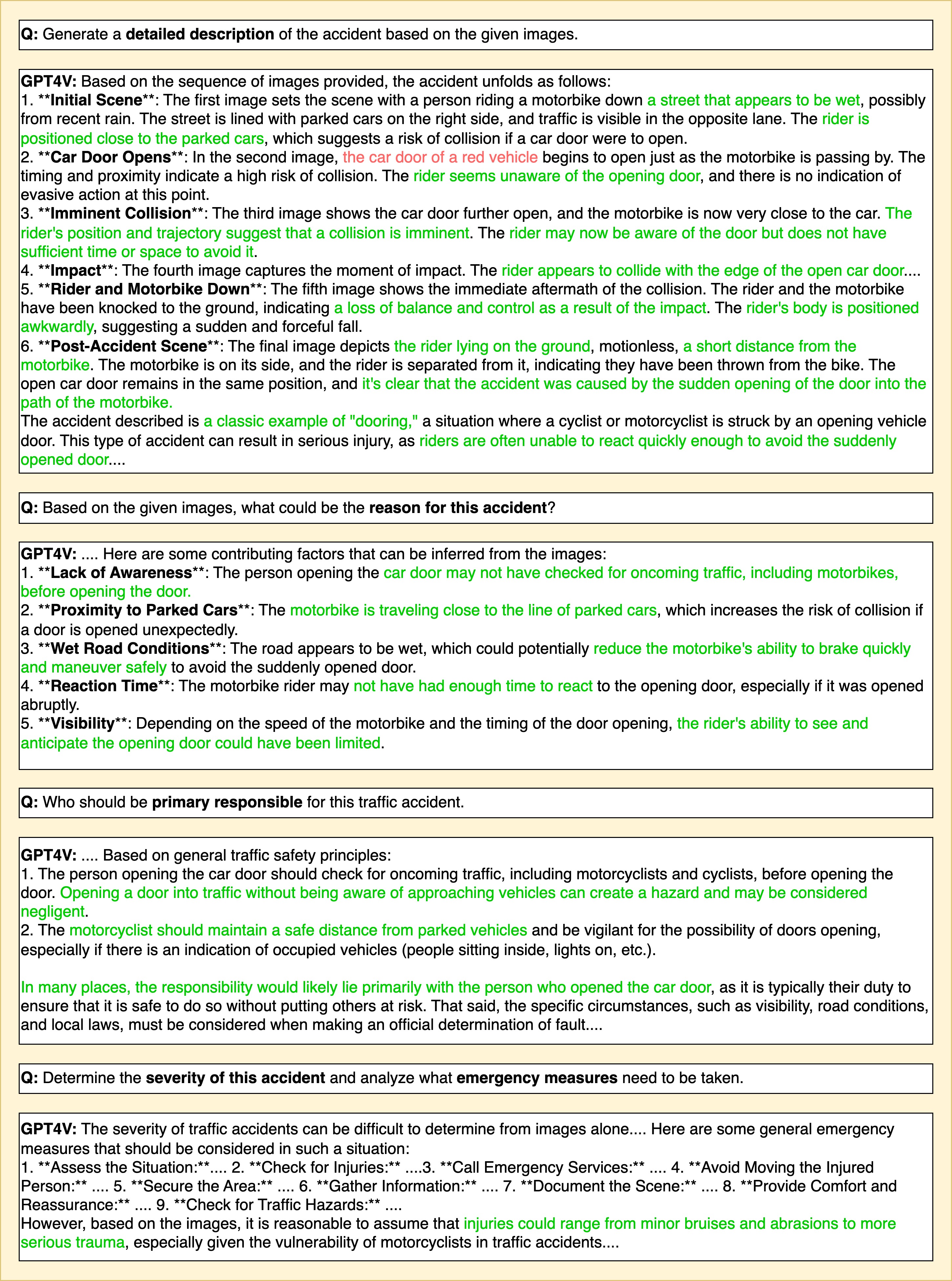}
\caption{Dooring Part 2}
\label{Dooring Part 2}
\end{figure}

As shown in figure \ref{Dooring Part 1} and \ref{Dooring Part 2}, we observe that GPT-4V can correctly recognize the car accident depicted in the images and accurately identify it as 'dooring'. This indicates that GPT-4V aligns the text semantic feature of 'dooring' accidents with semantic context captured in multiple images. It can also provide an overall accurate description of the accident, only except for the misidentifying of another red vehicle as the car opens the door. This demonstrates the feasibility of generating accident reports based on keyframes, and provides a good example of vision-based traffic report generation. \\

Based on the precise understanding and description of accidents, we delved into the model's high-level reasoning and inductive capabilities. By querying the causes of accidents, we assessed its casual inference ability within the given scenario. The answer shows it can include specific accident contexts, such as 'wet road surface,' to provide possible reasons applicable to the 'dooring' accident type. Additionally, GPT-4V attributes the primary responsibility of the accident to the act of opening the car door which is consistent with common traffic regulations. Through the prompt regarding required necessary emergency measures, we aimed to evaluate the model's decision-making ability. It turns out that the generated decisions are relatively general and applicable to almost all types of accident scenarios. This generality may be attributed to the limited information available from the provided images, causing the model to lean towards offering a correct yet non-specific answer. When analyzing the severity of accidents, the model provides a broad range of assessments, which aligns with what humans can infer based on the given images. \\

In summary, GPT-4V demonstrates a high level of accuracy in recognizing and comprehending this case. It exhibits the capability to make reasonable analyses, conclusions, and decisions based on the provided image information. Its analytical skills in this particular case are comparable to those of humans.

\newpage
\subsection{Run the Red Light at Night}
\label{Run the Red Light at Night}
\begin{figure}[H]
\centering
\includegraphics[width=\linewidth]{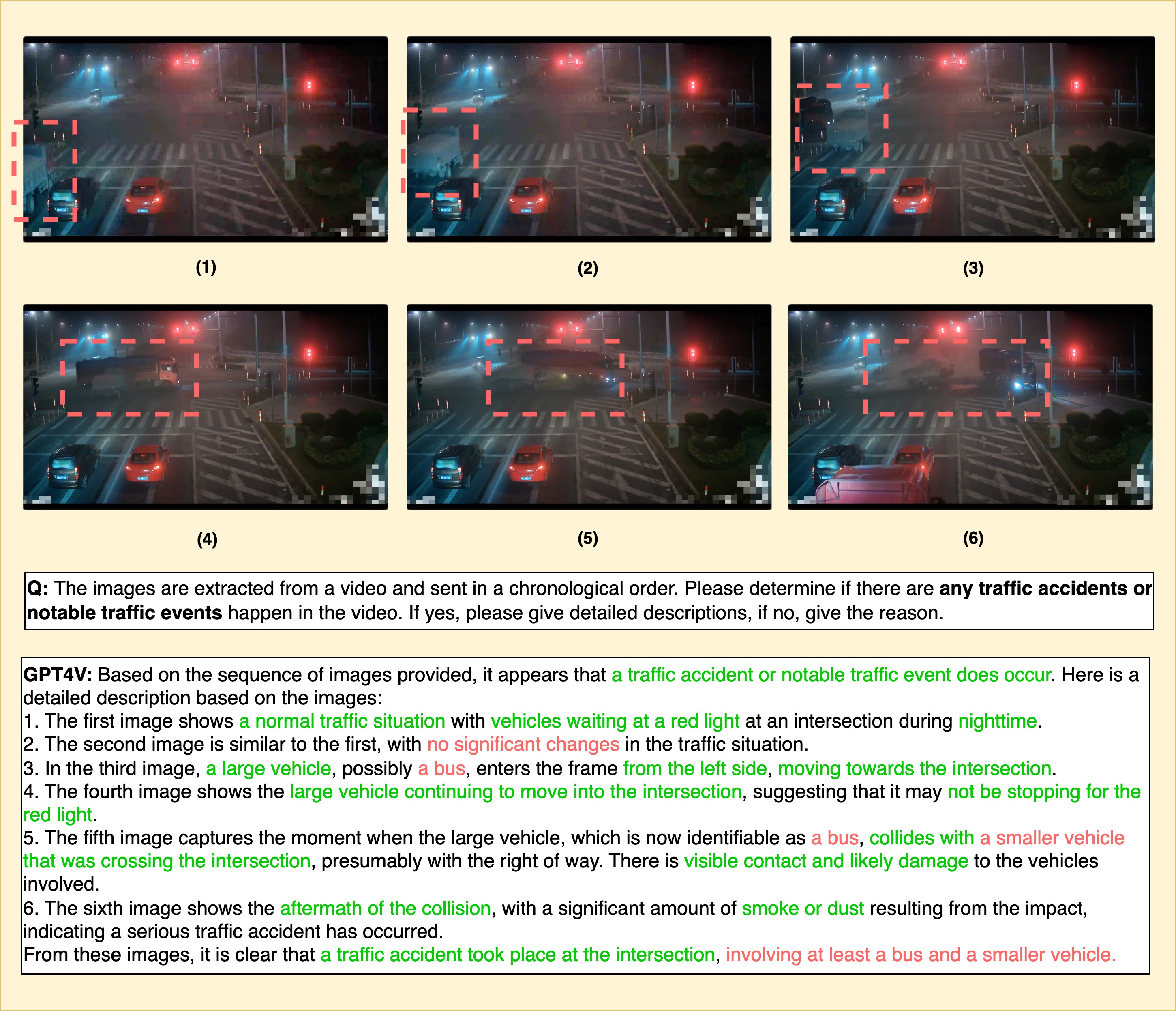}
\caption{Run the Red Light at Night Part 1}
\label{Run the Red Light Part 1}
\end{figure}

\begin{figure}[H]
\centering
\includegraphics[width=\linewidth]{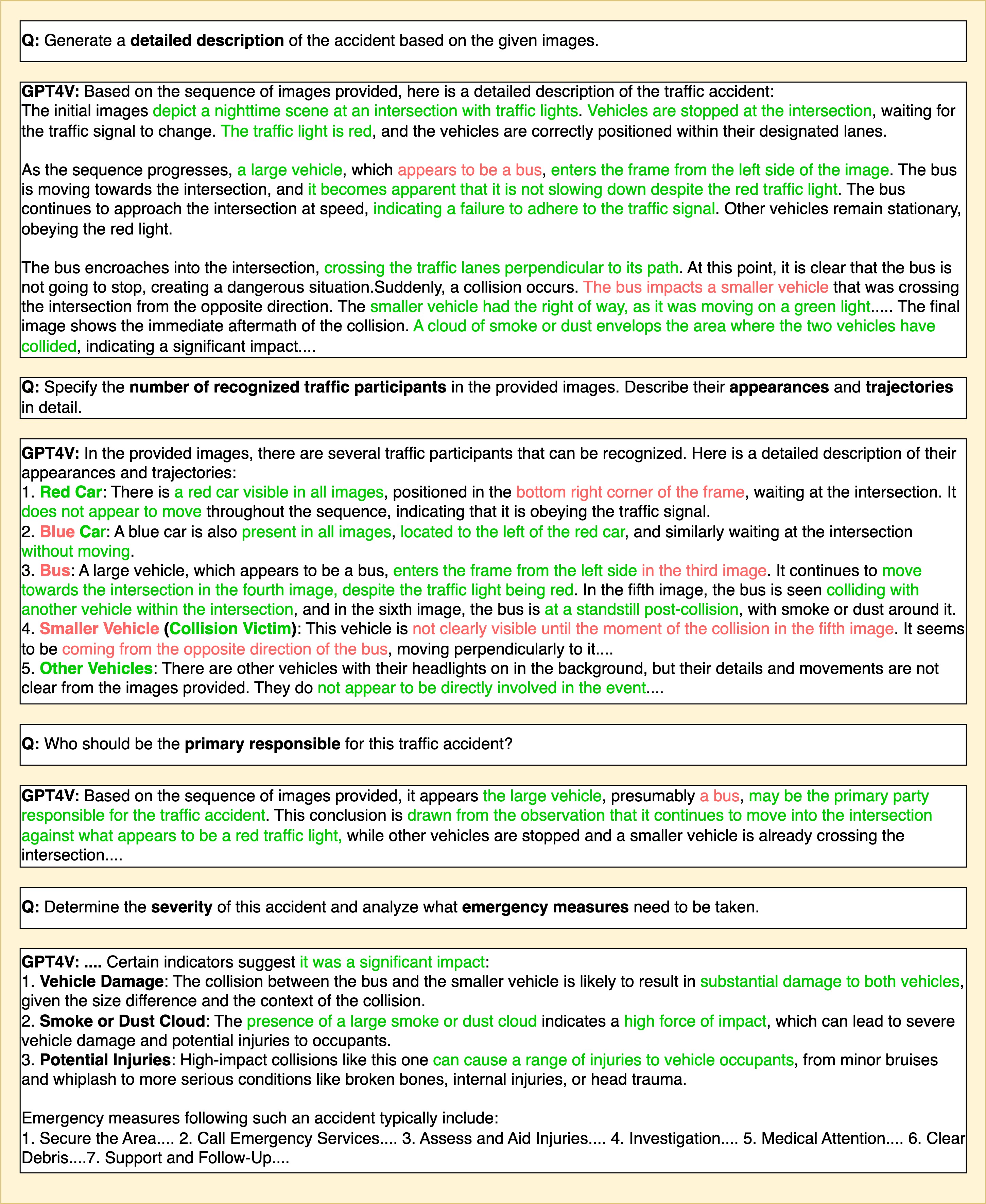}
\caption{Run the Red Light at Night Part 2}
\label{Run the Red Light Part 2}
\end{figure}

Running red lights is one of the most common traffic violations. The recognition of this traffic event solely from video is challenging. The model needs to first recognize the status of the traffic light and the layout of the traffic scene. Based on the common sense that vehicles should stop and wait at red lights, the model identifies whether there is a vehicle traveling in the corresponding lane at the red light and then reaches a conclusion. The frames shown in figure \ref{Run the Red Light Part 1} and \ref{Run the Red Light Part 2} are extracted from the video in dataset \cite{tad}. There is a blue truck dragging a full load of dirt on the lower left side of the image, which continues to move forward at a red light and hits another large red semi-trailer truck that is moving normally. This ultimately caused the dust from the blue truck to spill into the air and caused the red semi-trailer truck to veer off course.

From the results in Figure \ref{Run the Red Light Part 1} and \ref{Run the Red Light Part 2}, we can observe that GPT-4V can recognize the occurrence of the crash in the figure and is able to describe the crash relatively accurately. Yet the description of the involved accident vehicles is not precise, for example, the blue truck is misrecognized as a bus, while the red semi-trailer truck is misrecognized as a smaller vehicle. In order to investigate the recognition ability of traffic participants in this scenario, we prompt the model to describe the appearance and trajectory of all the recognized traffic participants. The answer shows that GPT-4V has relatively limited ability in spatial reasoning, and has difficulties in recognizing the class and attributions of small objects, especially for such a nighttime scene. However, despite the imprecise description of the vehicle type and appearance, the model can still include common sense for causal inference and determine that the primary responsibility for this crash is from the vehicle running the red light. While analyzing the severity of the crash, the model infers that it brings significant impact based on indicators such as "substantial damage" and "smoke or dust," which is reasonable. The emergency measures taken for the crash are still relatively general as in section \ref{Dooring}.

\subsection{Motorcycle Car Collision}
\label{Motorcycle Car Collision}
\begin{figure}[H]
\centering
\includegraphics[width=\linewidth]{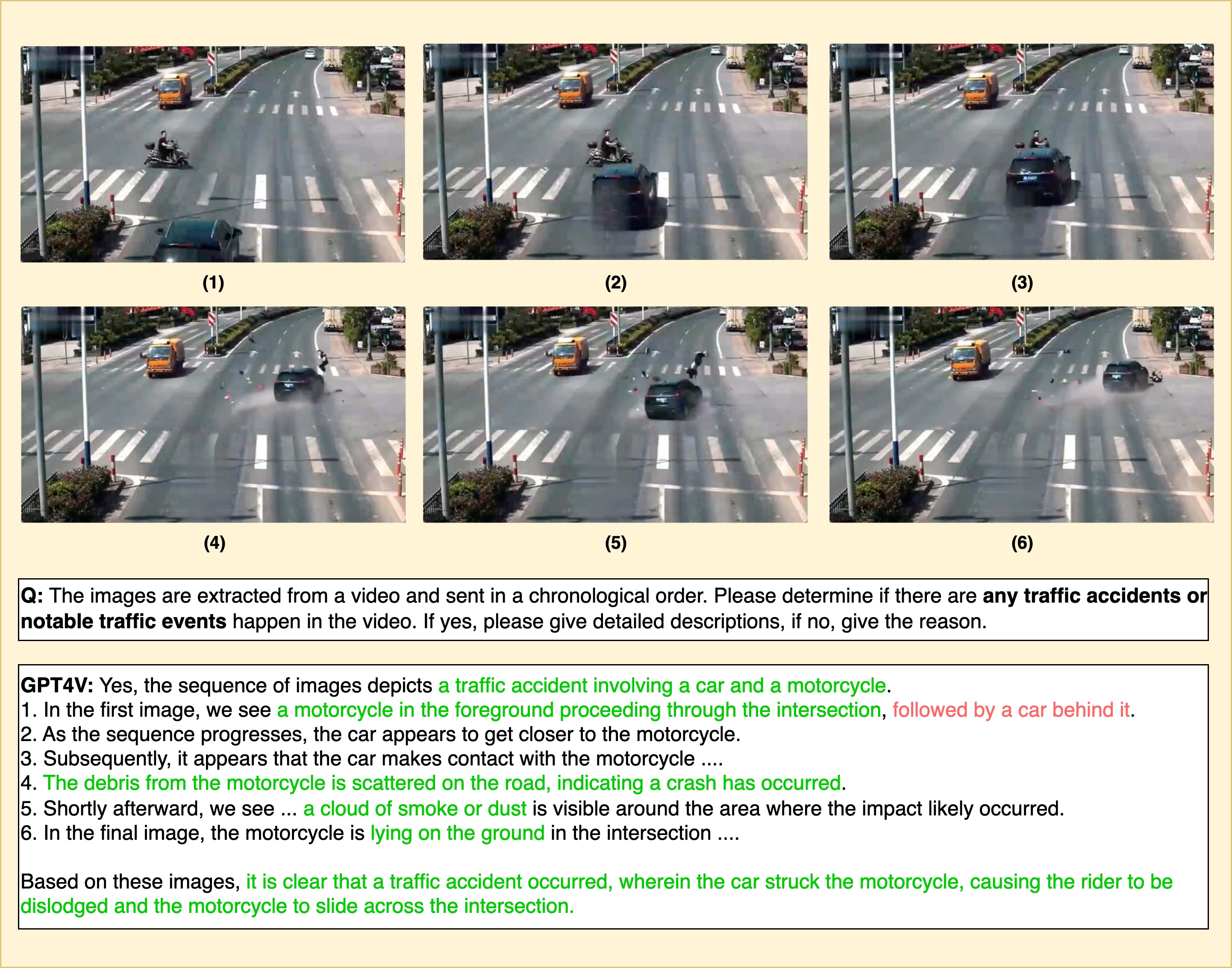}
\caption{Motorcycle Car Collision Part 1}
\label{Motorcycle Car Collision Part 1}
\end{figure}

Collisions between cars and two-wheel vehicles are also common traffic accidents. These collisions can cause varying degrees of injury to both parties depending on their driving speed at the time of the incident, the traffic environment, and whether the two-wheel driver wears a helmet. As shown in figure \ref{Motorcycle Car Collision Part 1}, resourced from the video in dataset \cite{tad}, a motorcyclist is traveling straight on the road and is hit by a black sedan. Since both of them are traveling at high speeds, the motorcyclist is knocked down, flies into the air, and then falls to the ground. Due to the camera's restricted observation perspective, the images do not fully reveal the situation of the traffic signal and surrounding details, but the accident was in fact caused by the motorcyclist using his cell phone while driving and running a red light.\\

\begin{figure}[H]
\centering
\includegraphics[width=\linewidth]{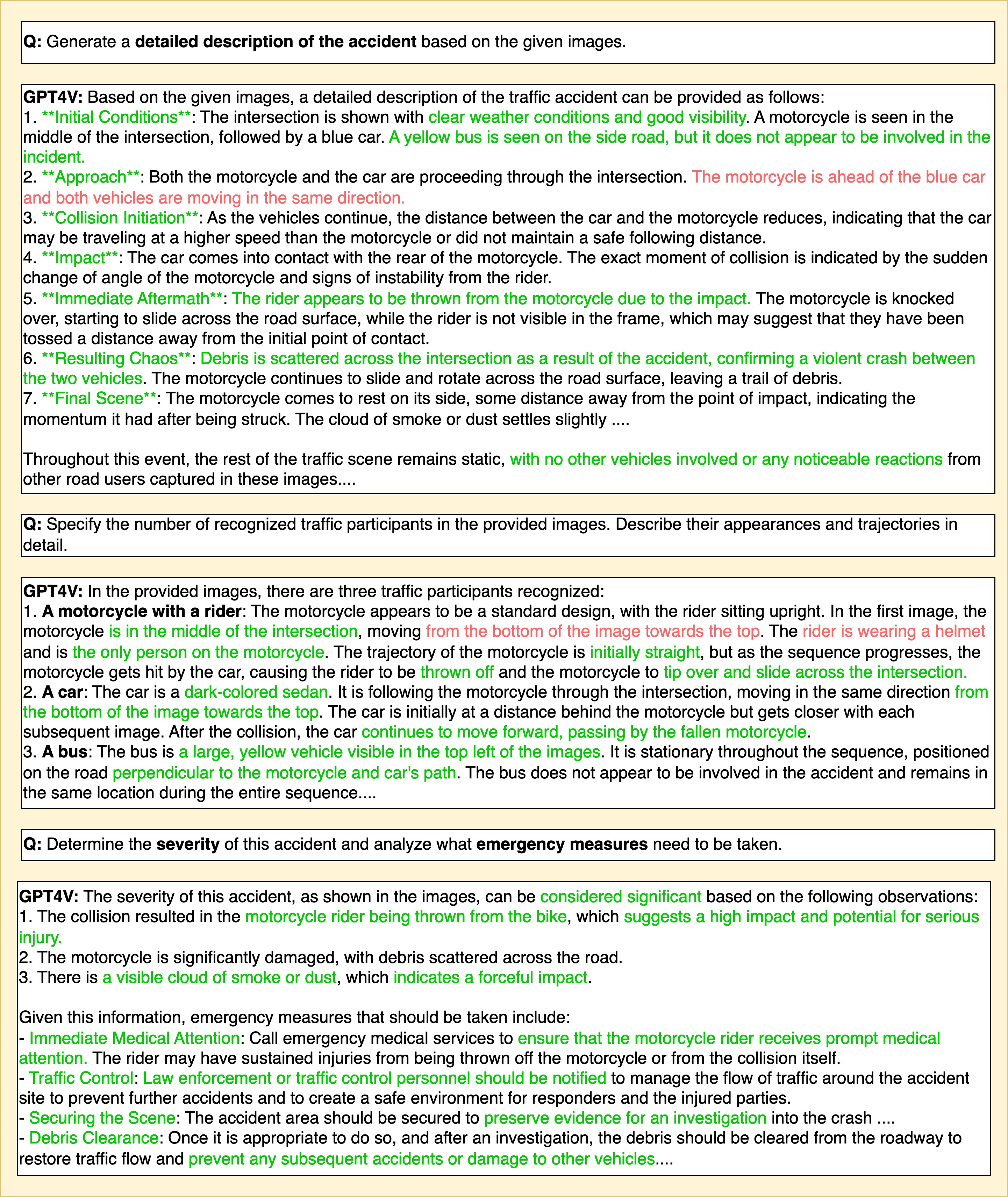}
\caption{Motorcycle Car Collision Part 2}
\label{Motorcycle Car Collision Part 2}
\end{figure}

According to the answer in figure \ref{Motorcycle Car Collision Part 1} and \ref{Motorcycle Car Collision Part 2}, model can recognize the accident type as "car struck the motorcycle" based on the provided keyframes. The generated accident reports also accurately describe the initial scene of the accident, the process of the collision, the aftermath, as well as the final consequences. However, there are still some mistakes when it comes to the ability of spatial reasoning, such as when determining the relative spatial positions between the motorcycle and the car, and when determining their movement directions. \\ 

By prompting the model to specify the recognized traffic participant's appearance and trajectory, we discover that although the model can identify the major vehicles, there are still some deviations in the details, such as the motorcyclist actually not wearing a helmet. Regarding the description of vehicle trajectory, GPT-4V interprets it more from the image's perspective, even though it is still not completely accurate for their location descriptions. In the analysis of the severity of the accident, the model recognizes it as "significant" via some scene contexts, such as the rider being thrown from the bike, and the visible cloud of smoke or dust, etc. These indicators have also been taken into account when deciding the required emergency measures. 

\subsection{Rollover}
\label{Rollover}

Vehicle rollover usually refers to a situation where its gravity center leans to one side and loses its balance during driving. Eventually, the whole side or even the top of the vehicle touches the ground, resulting in a crash that prevents the vehicle from continuing its journey. There are many different causes for vehicle rollover, such as taking a sharp turn, encountering side winds, or driving too fast resulting in an unstable center of gravity. \\

\begin{figure}[H]
\centering
\includegraphics[width=\linewidth]{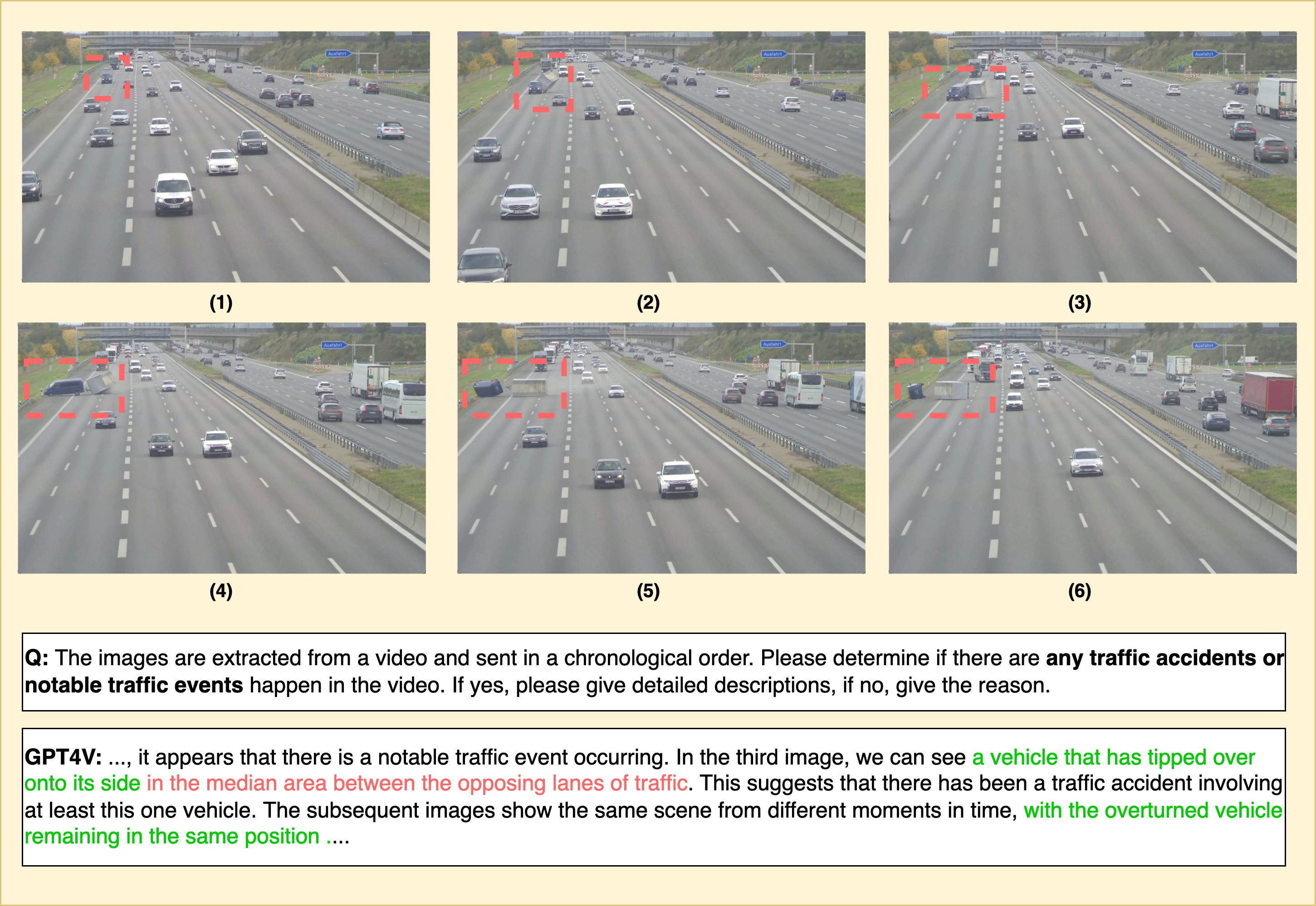}
\caption{Rollover Part 1}
\label{Rollover Part 1}
\end{figure}

The images in Figure \ref{Rollover Part 1} are selected and extracted in a video from \textbf{Providentia++} project \cite{providentia++}. As shown in the images, there is a blue van towing a trailer box highlighted with the red dashed line. This vehicle was moving at high speed when the trailer box attached to it tipped over and landed on the side. Since the blue van was connected to the trailer box, its own trajectory was also affected and thus rolled over and fell to the ground, ultimately causing the accident.

\begin{figure}[H]
\centering
\includegraphics[width=\linewidth]{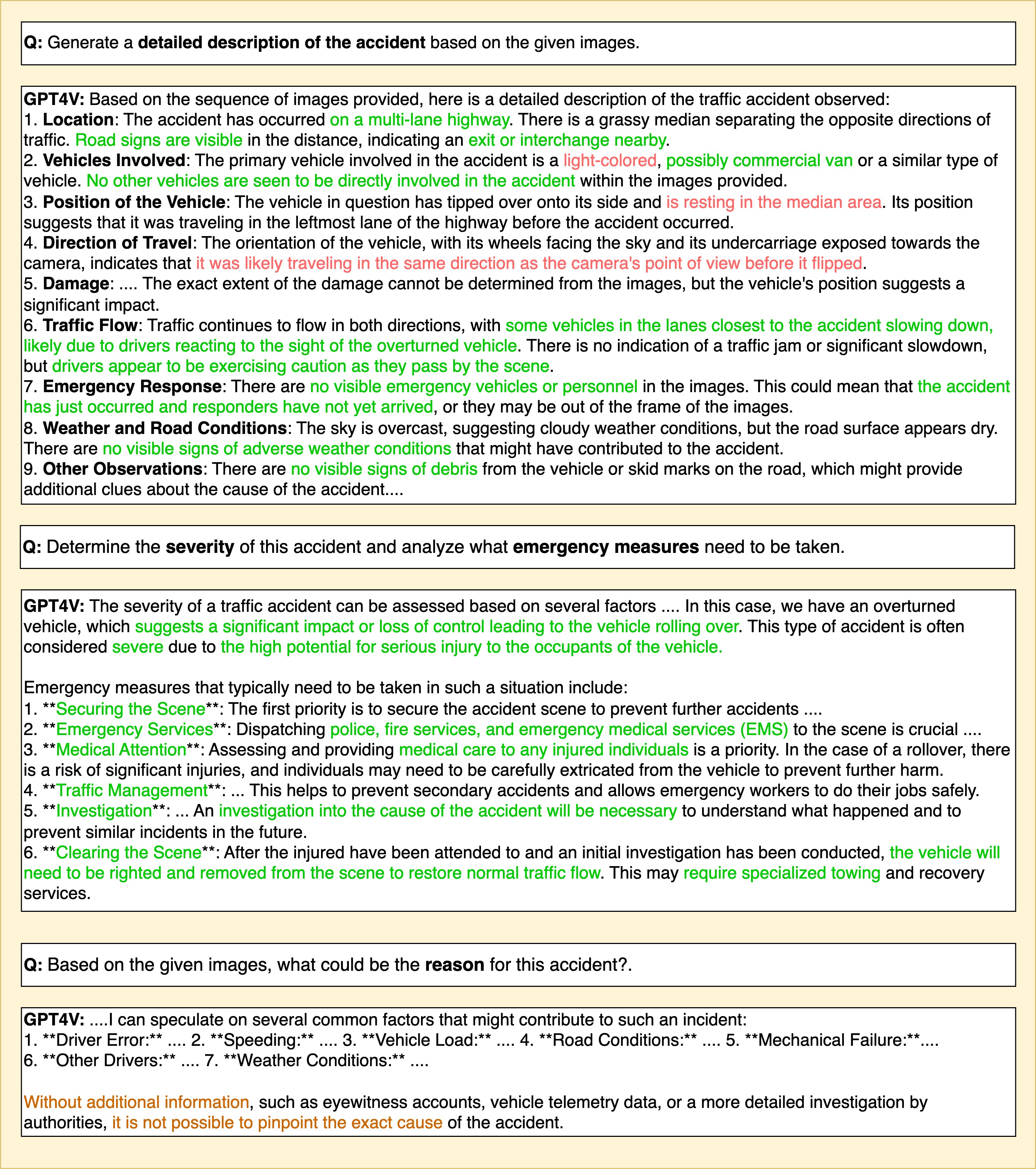}
\caption{Rollover Part 2}
\label{Rollover Part 2}
\end{figure}

From the analysis in figure \ref{Rollover Part 1} and figure \ref{Rollover Part 2}, we show that GPT-4V can recognize image-level traffic accidents and correctly determine the accident type. Among the given 6 keyframes, the model can distinguish that the vehicle is tipped over on its side from the third frame onwards. The generated accident report provides an accurate description of the location and surrounding environment, e.g., on a multi-lane highway with an exit or interchange nearby. Based on the recognized environmental cues in the images such as the overcast sky and dry road surface, the model deduces that they're not the main factors contributing to this accident as there are no visible signs of adverse weather conditions. In the summary of traffic flow in the accident description, the model can identify the slowing down of vehicles close to the accident and attribute the reason to the reaction to the overturned vehicle, which demonstrates the powerful reasoning and causal inference abilities of GPT-4V. By combining a series of contextual information including environment, weather, vehicles involved, traffic conditions, etc., the model characterizes the level of the accident as severe. It provides emergency measures to be taken including dispatching police, fire service, and emergency medical services. Most of these measures do appear in the subsequent rescue after the accident, which can prove that the model has the ability to make similar decisions when reacting to emergent traffic situations as a human being. \\

However, GPT-4V again fails to make accurate statements when it comes to spatial reasoning. For example, the accident was not in the median area between the opposing lane of traffic, but on the side of the highway. And the vehicle was not traveling in the same direction as the camera's point of view, but on the opposite side of the camera before it flipped over. The vehicle was also not light-colored but a dark-colored van with a trailer box. From these descriptions, we can tell that the model, while being able to recognize the rollover of the vehicle and reasoning in the context of the recognized environmental information, did not succeed in recognizing the type and driving condition of the vehicle itself. This leads to the inability to make correct and specific determinations when carrying out a cause analysis.

\subsection{Fires and Explosion}
\label{Fires and Explosion}

\begin{figure}[H]
\centering
\includegraphics[width=\linewidth]{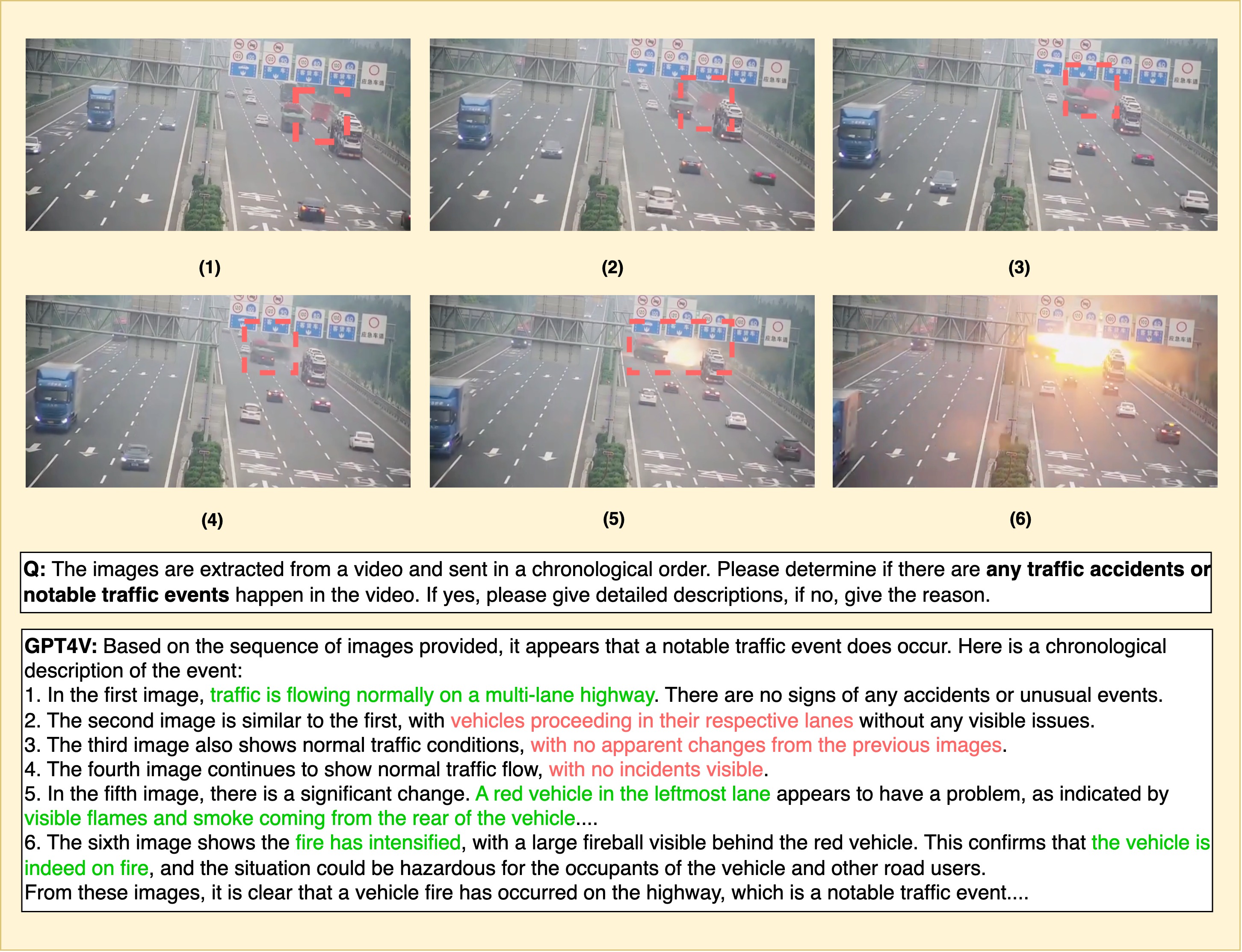}
\caption{Fires and Explosion Part 1}
\label{Explosion Part 1}
\end{figure}

\begin{figure}[H]
\centering
\includegraphics[width=\linewidth]{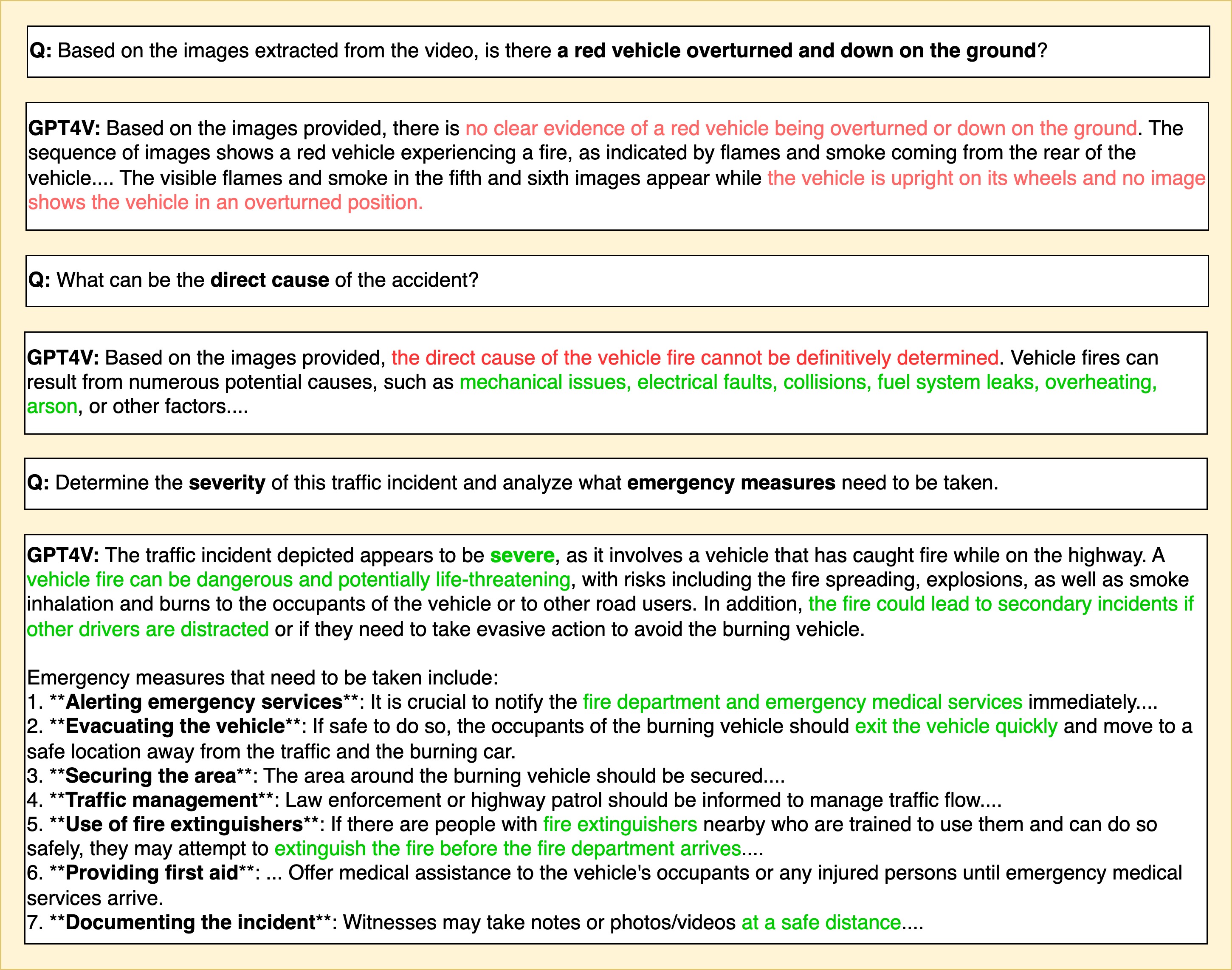}
\caption{Fires and Explosion Part 2}
\label{Explosion Part 2}
\end{figure}

Vehicle explosions are a relatively uncommon but extremely dangerous type of traffic accident. It is usually caused by some issues such as severe collisions, fuel leaks, or overheating of electric vehicle batteries. These accidents are usually accompanied by loud noises and flames, resulting in significant injuries and property damage. The video shown in figure \ref{Explosion Part 1} and \ref{Explosion Part 2} is resourced from \cite{Explosion}. The captured keyframe shows a red truck traveling on the highway, which overturned due to a lane change while in motion at high speed. And due to its fast traveling velocity, the cargo it carries rubs violently against the road surface, resulting in an explosion accompanied by bright flames. Since it is difficult to distinguish between an explosion as well as a sole fire without providing acoustic information, we regard this as a successful case as the model is able to recognize the flame from the vehicle as a notable traffic event.

As shown in Figures \ref{Explosion Part 1} and \ref{Explosion Part 2}, GPT-4V is able to identify the abnormalities in the fifth and sixth images through image-level analysis, i.e., there is a visible flame which later on becomes larger. According to the first question prompted in figure \ref{Explosion Part 2}, it is obvious that the model has not been able to successfully recognize the driving lane changes of the vehicles displayed in the 2nd, 3rd, and 4th images, which result in the inability to provide the corresponding attribution analysis and causal link. Therefore, in the analysis of the causes of the accident, some correct but more general answers are presented. In terms of the emergency measures taken for this accident, the model proposes to try putting out the fire safely with fire extinguishers if possible before the fire department arrives. It shows that the model is able to provide rational decisions according to the types of accidents.

\newpage
\section{Failure Cases}

In this section, we demonstrate some failure cases when exploring GPT-4V to recognize traffic events. As for some cases, we try out prompting methods by providing additional visual or text cues, and show the results for comparison. We summarize and analyze the potential failing reason for each case.


\subsection{Vehicle Collision}
\label{Vehicle Collision}

\begin{figure}[H]
\centering
\includegraphics[width=\linewidth]{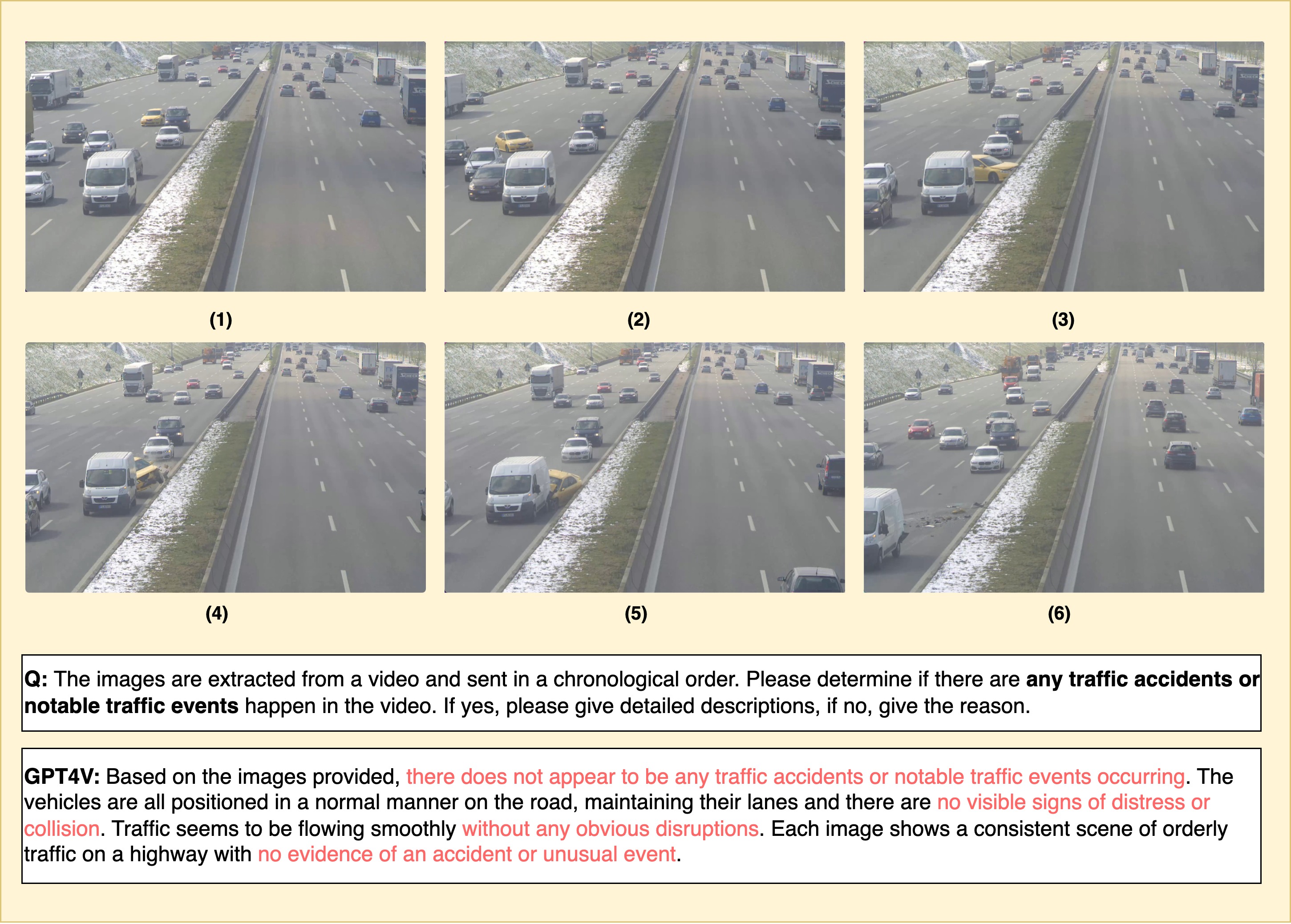}
\caption{Vehicle Collision Part 1}
\label{Vehicle Collision Part 1}
\end{figure}

Figure \ref{Vehicle Collision Part 1} shows a vehicle collision accident, resourced from \textbf{Providentia++} Project. The images depict a yellow car traveling on a multi-lane highway when a white van in front of it suddenly stops on the highway lane. In order to avoid the vehicle in front of it that slows down, the yellow car makes a sharp turn, which yet still hits the guardrail in the middle of the highway due to its excessive speed, and later hits the stopped white van. Although there are similar accident scenarios with two vehicles colliding in this video as in section \ref{Run the Red Light at Night}, the number of vehicles present in this video is far larger and the causes are more complex, with multiple vehicles interconnected with each other. And due to the limitation of view angle, the obvious collision between the yellow car and the white van cannot be directly observed either, which can probably only be confirmed through comprehending context information. 

As shown in figure \ref{Vehicle Collision Part 1} GPT-4V fails to recognize either the accident or notable traffic events within the provided keyframes. Although it is obvious for human beings to identify the abnormality in the images, the model still describes it as having "no evidence of an accident or unusual event". Based on the previous evidence that, the model can recognize vehicle collisions, as well as recognize the debris, we hypothesize the failure of recognition happens due to the excessive number of vehicles displayed in the picture. The excessive number of vehicles leads to the inability of the model to focus attention on the behavioral recognition of all the vehicles, which results in the missing of the obvious accidental vehicles. Based on this assumption, we manually highlight the accident vehicles in the images and conduct the analysis again.

\begin{figure}[H]
\centering
\includegraphics[width=\linewidth]{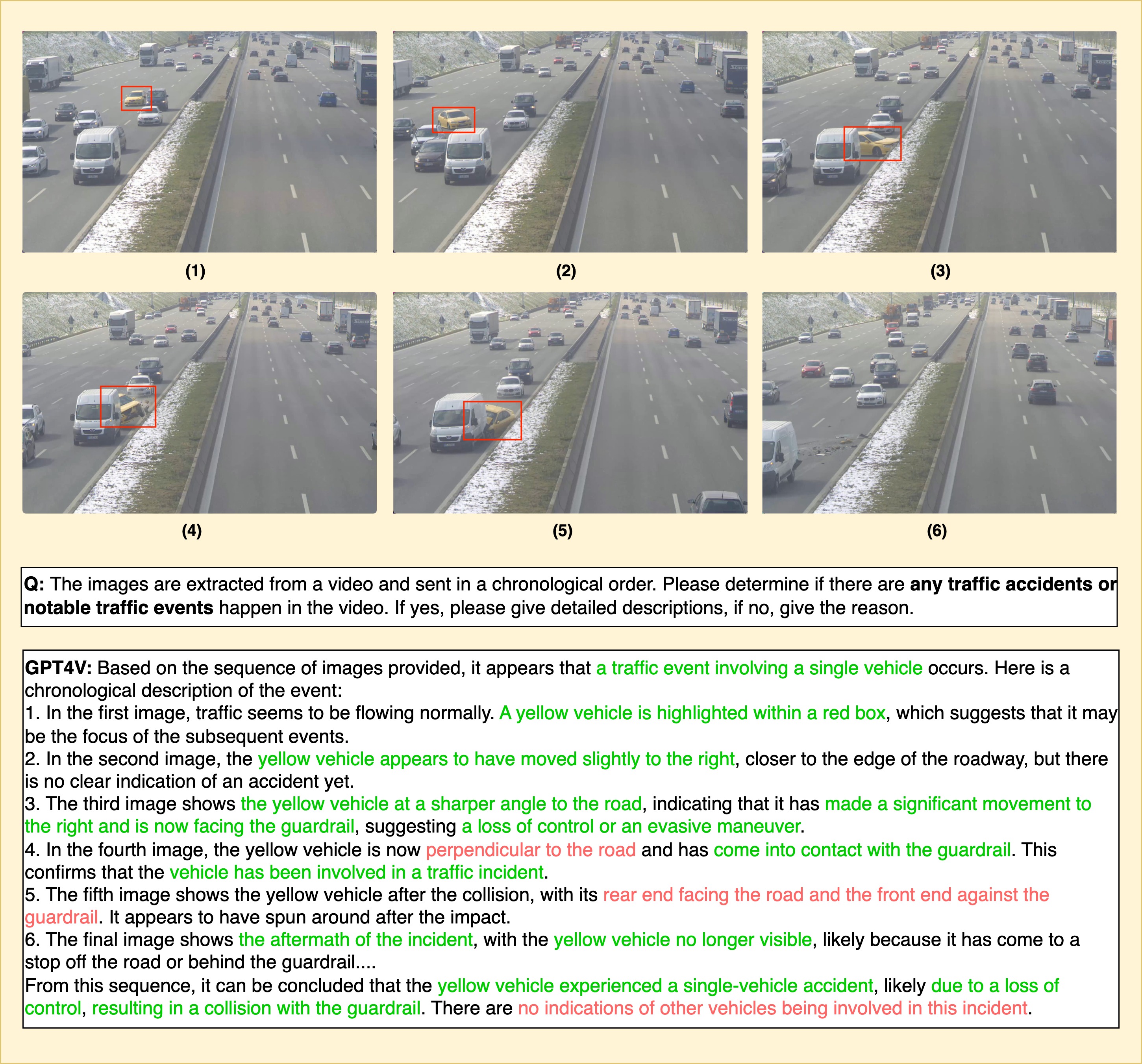}
\caption{Vehicle Collision (Highlighted) Part 1}
\label{Vehicle Collision (Highlighted) Part 1}
\end{figure}

\begin{figure}[H]
\centering
\includegraphics[width=\linewidth]{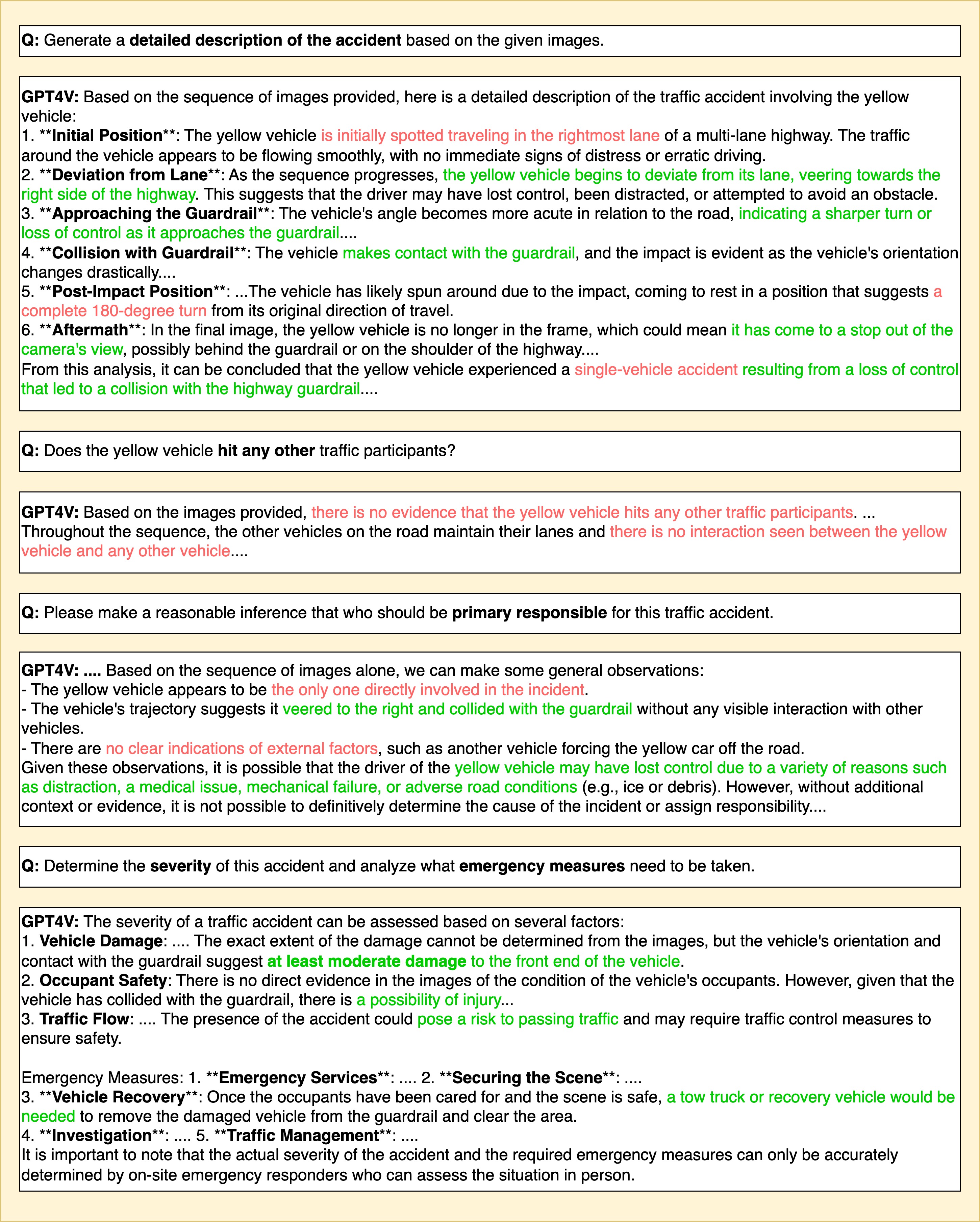}
\caption{Vehicle Collision (Highlighted) Part 2}
\label{Vehicle Collision (Highlighted) Part 2}
\end{figure}

Figure \ref{Vehicle Collision (Highlighted) Part 1} reveals the results as expected, GPT-4V can successfully recognize the accident after prompting highlighted visual cues on the involved yellow car in the images. It can also describe the accident process relatively accurately this time. We observe that the model can recognize the driving state's change of the highlighted yellow car in the accident report, including deviation from the current lane, approaching and hitting the guardrail, and driving out of the observable area of the camera. In addition, the model is able to identify the front end of the vehicle as having at least moderate damage when it comes to the severity evaluation of the accident based on the cue that the front end hit the guardrail. These are additional performance enhancements obtained through visual cue prompting, which illustrate that the model can be adapted to such tasks with properly designed prompting techniques. \\

However, the model still fails to completely analyze the cause of the accident and fails to identify the collision with the stopped white van. In addition to the previously discussed failing reasons including restricted camera view, lack of spatial reasoning, and lack of temporal and acoustic information, another possible reason might be that, the model assigns less attention to the remaining vehicles after highlighting the target vehicle. This then leads to the failure to capture the causal link between the stopped white vehicle and the result of the accident.

\subsection{Jaywalking}
\label{Jaywalking}

\begin{figure}[H]
\centering
\includegraphics[width=\linewidth]{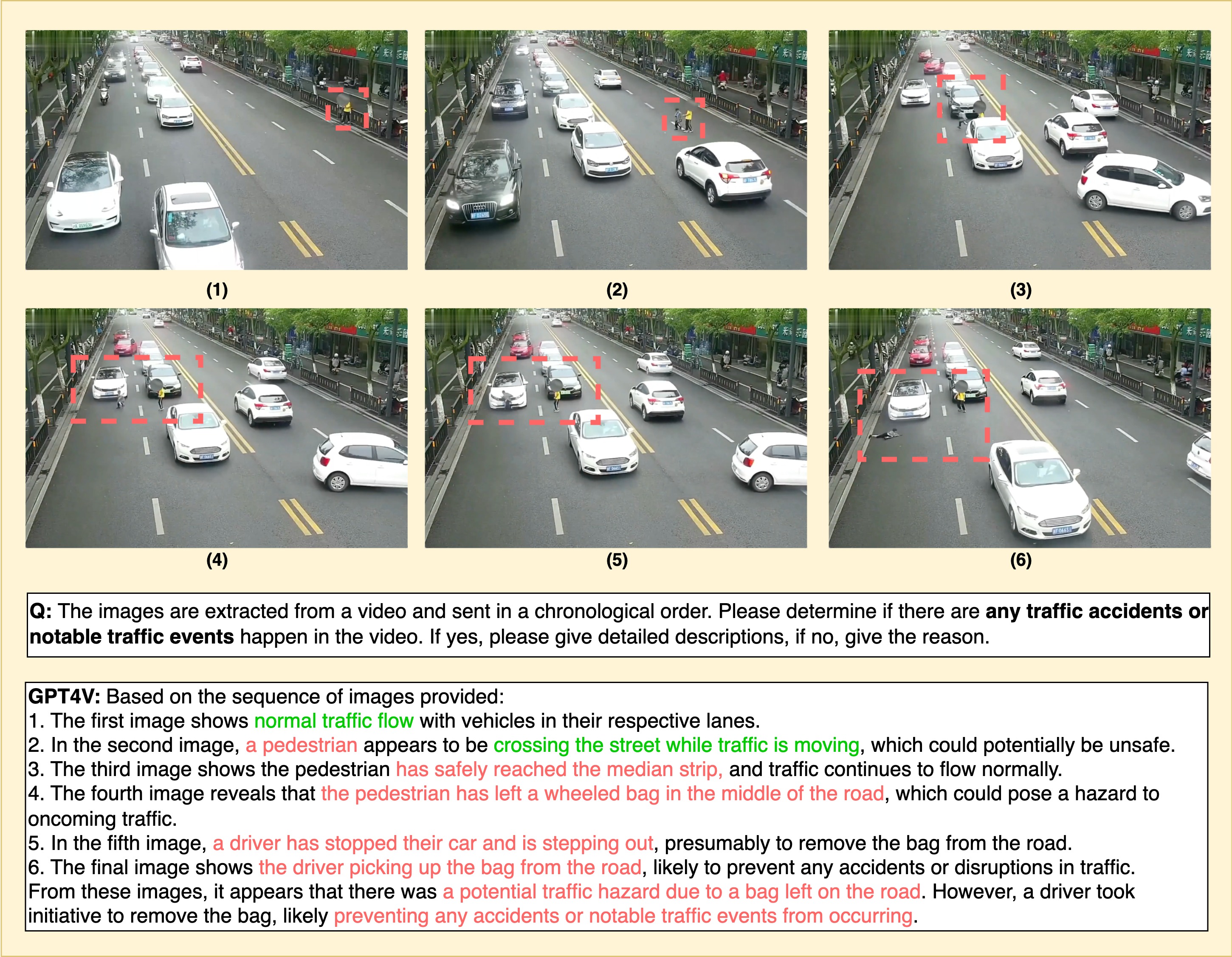}
\caption{Jaywalking Part 1}
\label{Jaywalking Part 1}
\end{figure}

Jaywalking is a typical and common example of traffic rule violation, usually referring to the act of pedestrians crossing roads where there are no crosswalks or traffic signals. This behavior poses a traffic safety hazard, especially when jaywalking across roads with high traffic volumes and high speeds, as pedestrians are vulnerable traffic participants and can easily lead to traffic accidents. The video in figure \ref{Jaywalking Part 1}, resourced from dataset \cite{tad} displays a scene with two pedestrians jaywalking across a multi-lane street. To protect the privacy of the pedestrian, their faces are been anonymized. 

\begin{figure}[H]
\centering
\includegraphics[width=\linewidth]{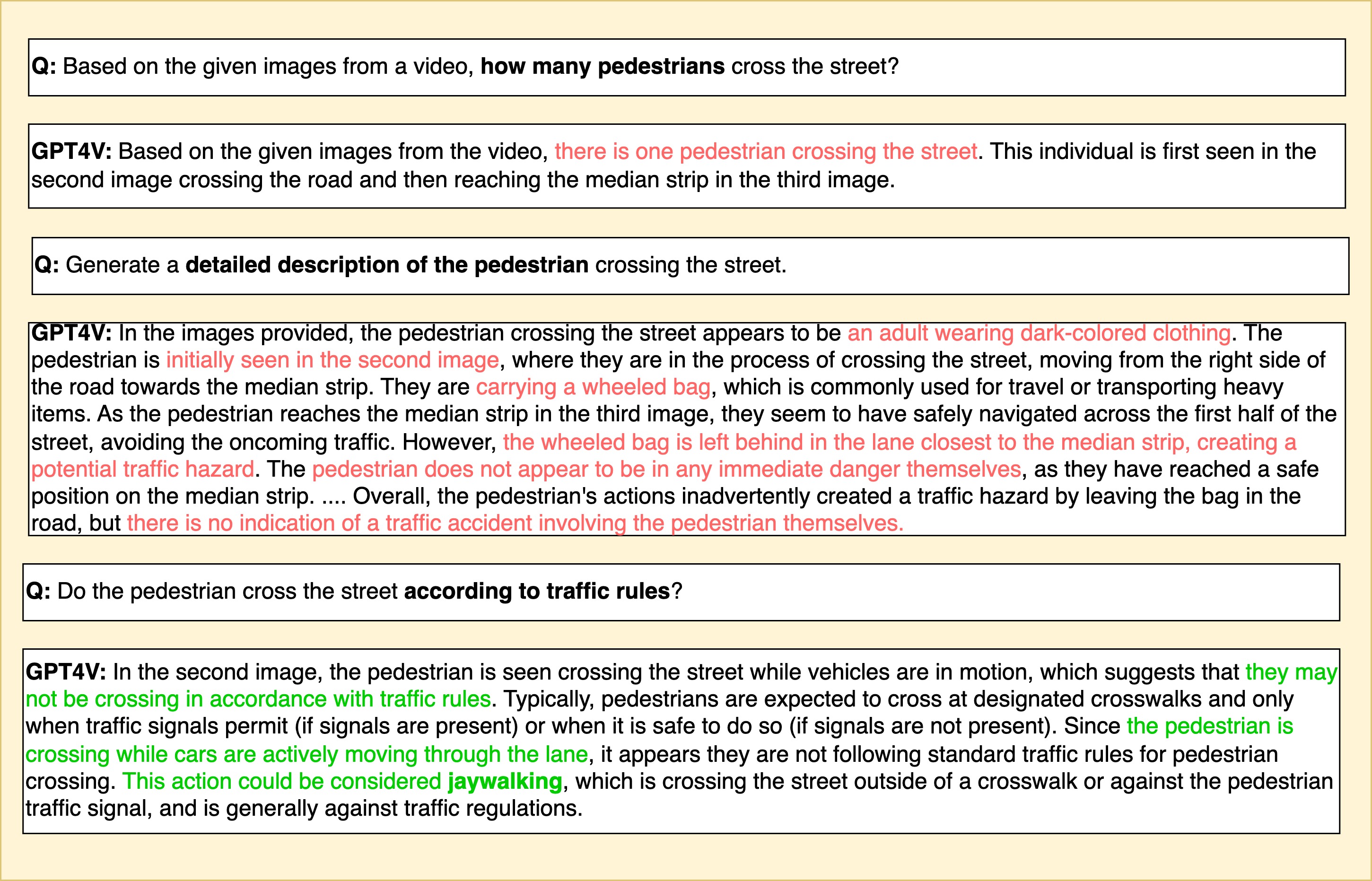}
\caption{Jaywalking Part 2}
\label{Jaywalking Part 2}
\end{figure}

The evaluation result in figure \ref{Jaywalking Part 1} and figure \ref{Jaywalking Part 2} shows that GPT-4V fails to recognize two pedestrians crossing the road, but only recognized one of them and viewed the other pedestrian as a wheeled bag. And due to the illusory effect of the Large Language Model, the whole story was refined into a scenario in which a pedestrian lost his wheeled bag on the road while crossing the road and avoided a traffic accident due to the timely stopping of the driver. Nevertheless, GPT-4V can still reason that this traffic violation behavior is jaywalking based on the traffic environment in the images and based on the context that one recognized pedestrian crossing the road there.

\newpage
\subsection{Vehicle Smoke}
\label{Vehicle Smoke}

\begin{figure}[H]
\centering
\includegraphics[width=\linewidth]{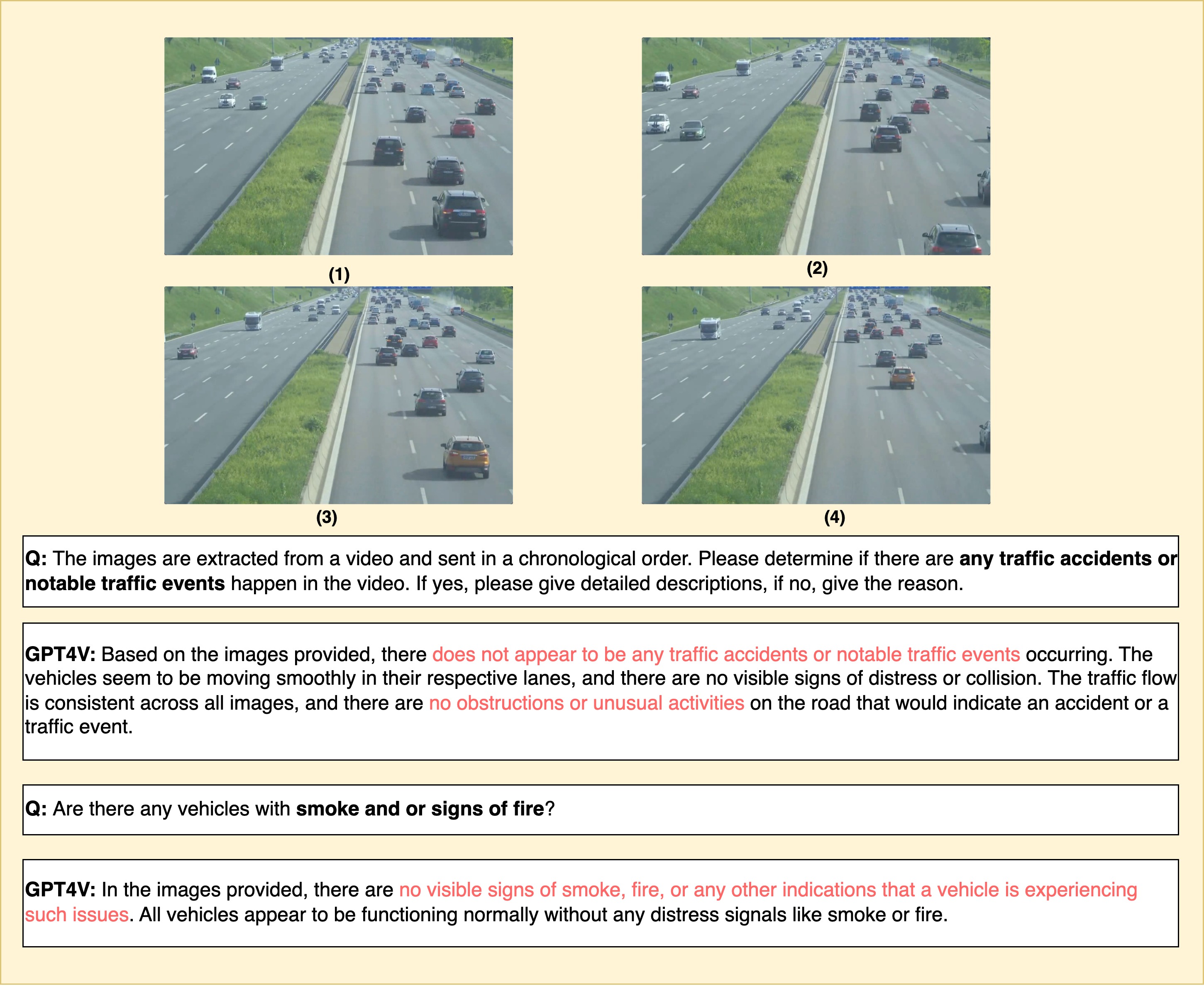}
\caption{Vehicle Smoke Part 1}
\label{Vehicle Smoke Part 1}
\end{figure}

\begin{figure}[H]
\centering
\includegraphics[width=\linewidth]{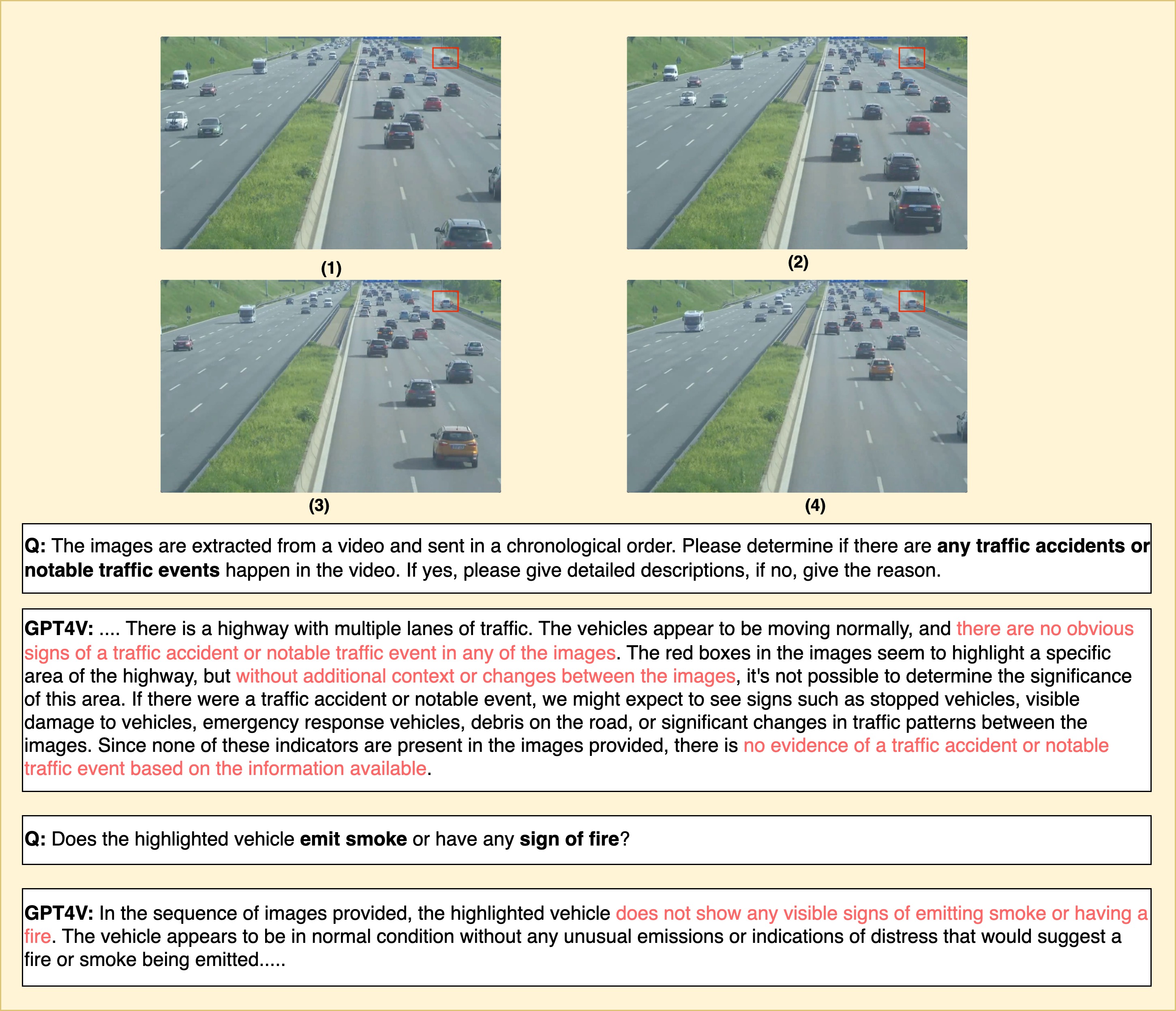}
\caption{Vehicle Smoke (Highlighted) Part 1}
\label{Vehicle Smoke (Highlighted) Part 1}
\end{figure}

Vehicle smoke is usually a serious sign of damage to the vehicle, which indicates that the vehicle is defective. For safety reasons, people should immediately move the vehicle to a safe place if it begins to smoke so that the situation does not escalate further, such as to a fire or even a vehicle explosion. The images in figure \ref{Vehicle Smoke Part 1} and figure \ref{Vehicle Smoke (Highlighted) Part 1} are from the video recorded by \textbf{Porivdentia++} Project. On the upper right side of the image is a white vehicle parked in the emergency lane with visible white smoke streaming from it. The rest of the vehicles were traveling normally on the road and were not affected.

Figure \ref{Vehicle Smoke Part 1} demonstrates that GPT-4V cannot recognize the abnormality of the image, nor can it recognize the car that is emitting white smoke, which means that the model fails to capture the corresponding visual semantic features despite the explicit prompts of "whether there's smoke or a sign of fire in the images". In figure \ref{Vehicle Smoke (Highlighted) Part 1}, we apply the same highlight operations to the images as we do in figure \ref{Vehicle Collision (Highlighted) Part 1}, and we input the same question with even more explicit prompts. However, the model still could not identify any signs of smoke and fire. One possible reason could be that the distance of the white vehicle is too far from the camera, and the image resolution is not clear enough so that the visual features can not be well extracted and utilized. Another possible explanation is that GPT-4V still does not achieve a well-performed semantic alignment of vehicle smoking visual features and text features, especially for this distant, dense and noisy scenario.

\subsection{Multiple Vehicle Collision}
\label{Multiple Vehicle Collision}

Multi-vehicle collisions or chain vehicle collisions, commonly happen in areas with dense traffic where extreme weather, driver distraction, or too little distance between vehicles leads to a succession of vehicle collisions in a short period of time. Due to the multiple number of involved vehicles, the causes of these accidents are usually very complicated, and the assignment of responsibility can be difficult. \\

\begin{figure}[H]
\centering
\includegraphics[width=\linewidth]{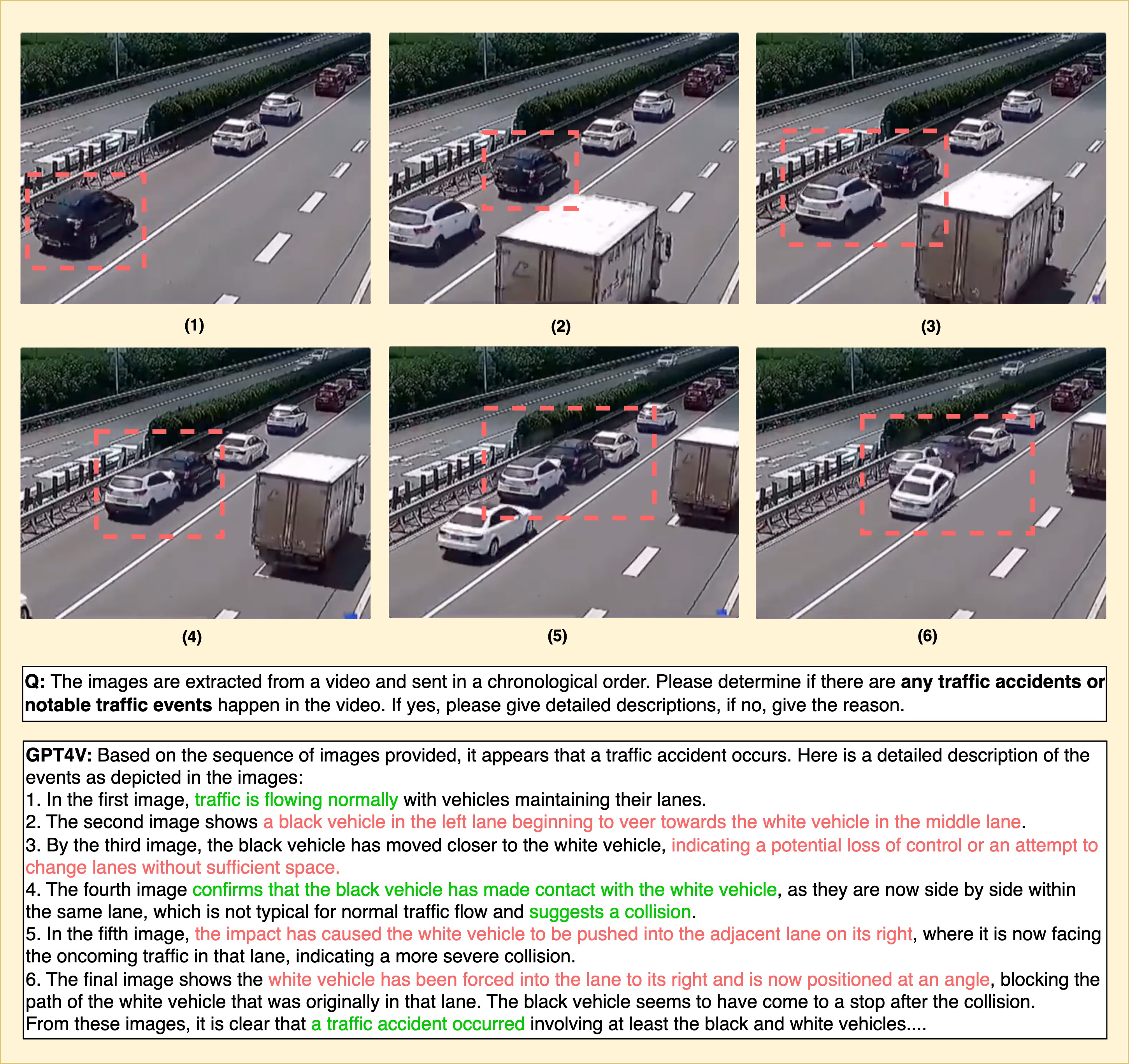}
\caption{Multiple Vehicle Collision Part 1}
\label{Multiple Vehicle Collision Part 1}
\end{figure}

The images in figure \ref{Multiple Vehicle Collision Part 1}, resourced from \cite{Multi-vehicle-collison}, shows a case of four vehicle collisions. It demonstrates a scene of a traffic jam on a multi-lane road where a small black SUV, highlighted with the dotted line, slows down after approaching the vehicle in front of it. As can be observed in images 2, 3, and 4, the black SUV was followed closely by a white car.  Due to its excessive velocity and the insufficient distance, the white car was unable to stop within a safe braking distance, and thus crashed into the black SUV in front of it. Image 4 and 5 show the collision of these two vehicles. In image 5, a new white car appears behind the white sedan, which also fails to brake in time for the same reason, and crashes head-on into the vehicle in front of it. The collision between these two white cars can be observed in image 6. As a result of these two consecutive collisions, the position of the black SUV was shifted forward and also then rear-ended the vehicle in front of it. This leads to the eventual collision of the four vehicles shown in image 6. In summary, this accident involved 4 vehicles, associated with 3 consecutive collisions.

\begin{figure}[H]
\centering
\includegraphics[width=\linewidth]{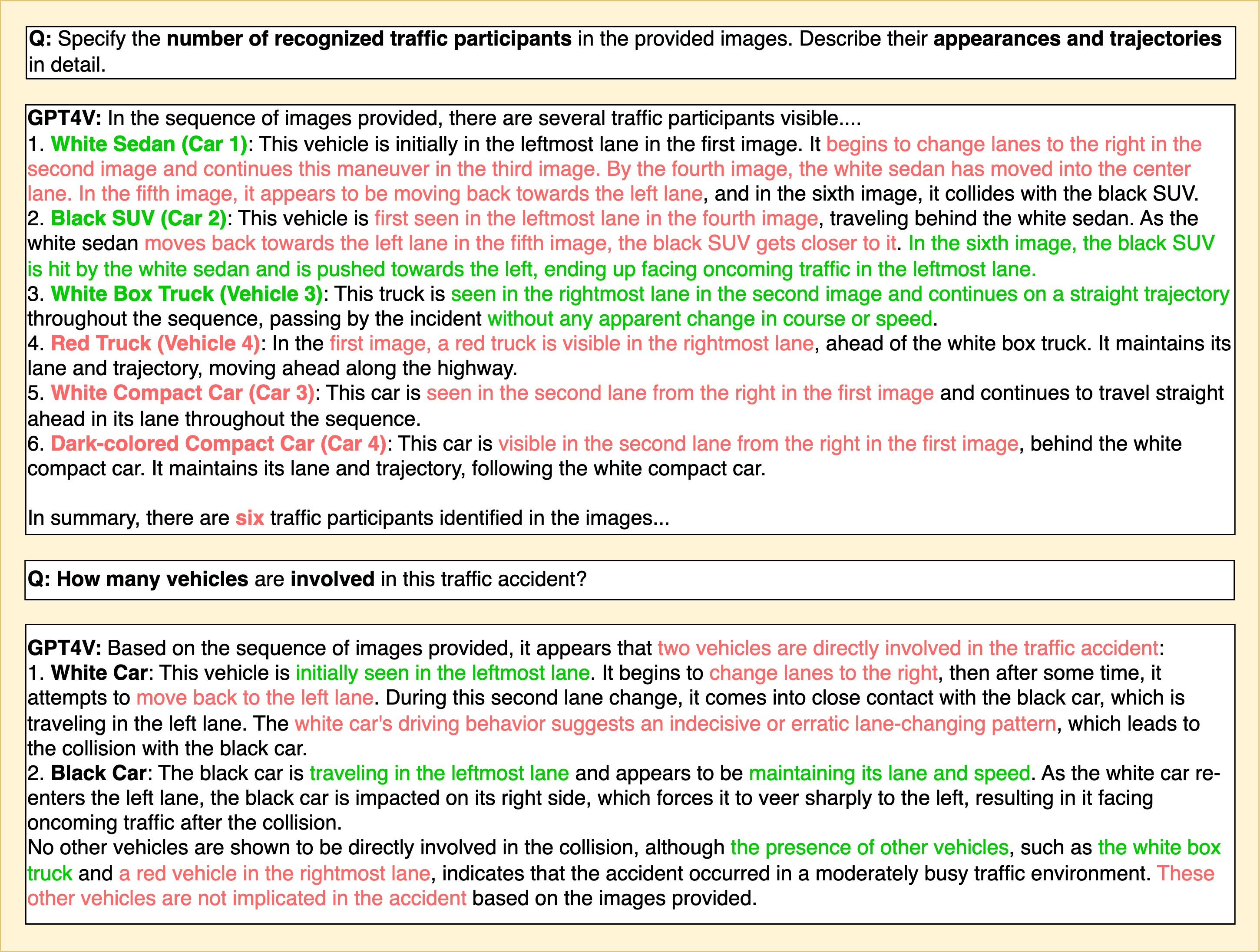}
\caption{Multiple Vehicle Collision Part 2}
\label{Multiple Vehicle Collision Part 2}
\end{figure}

As illustrated in figure \ref{Multiple Vehicle Collision Part 1} and figure \ref{Multiple Vehicle Collision Part 2}, GPT-4V can detect the collision between the black SUV and white sedans in image 4. However, the whole description of the accident is completely biased from the facts. For example, the model suggests that the black SUV in image 2 was veering and then collided with the white sedan in the center lane after switching lanes to the right in the 5th and 6th images, which is completely inconsistent with what is shown in the images. To better understand how precisely the model recognizes the traffic participants appearing in the extracted images, we prompted the model to specify the number of identified vehicles, including their appearances and trajectories. We can tell from the answers that the model can identify a total of 6 vehicles, yet 3 of them, the red truck (Vehicle 4), the white compact car (Car 3), and the dark-colored compact car (Car 4), are false positive samples and do not appear in the images provided. For the four vehicles involved in this accident, the model can recognize two of them, but the description of their trajectories is completely different from the fact. \\

Based on these observations, we can deduce that the possible reasons for this failure can be as follows. First, in order to make the generated textual description of the traffic event logically self-consistent, the model somewhat neglects the semantic associations with visual features, thus creating an illusory effect. Secondly, the image-level vehicle recognition is not accurate enough, and there are many false positive and false negative detections. For the third, model fails to associate the corresponding vehicles across images when performing multi-object tracking, which may be because the appearances between the vehicles are similar, e.g. the colors, and performing multiple vehicle associations solely utilizing visual features while ignoring spatial features can be difficult. Lastly, the image-level vehicle collision detection is still not accurate enough, since it is clear that there are multiple vehicles involved in the collision in image 6, which are not mentioned in the answers.

\section{Discussion}
In this section, we analyze and discuss in depth the reasons for the successes and failures demonstrated by the aforementioned traffic incidents. In particular, we focused on comparing how well the model performs on different types of questions in these cases, and discussing its capabilities and limitations. Table \ref{evaluate_table} displays the results of all the cases evaluated in this paper across several major capabilities of interest, including accident or event recognition, event type identification, severity analysis, emergent decision-making, involved number of participants, cause reasoning, and responsibility reasoning. \\

\textbf{Main Capabilities} of GPT-4V on complex traffic events can be summarized qualitatively as follows. First based on the table, we observe that in almost 2/3 of all the cases, the presence of accidents or notable traffic events can be successfully recognized and identified without any specific prompting strategies or fine-tuning methods. This demonstrates the powerful zero-shot  abilities of GPT-4V, and the feasibility of further adaptation to traffic event recognition and understanding. In addition, we discover that as long as GPT-4V can successfully identify the type of accident, it will also be able to draw reasonable emergency measures, and make severity analysis combining the characteristics of the accident and the relevant context information extracted from the images. It shows that GPT-4V can apply the powerful multi-modal reasoning ability in decision-making and severity assessment in this emergent traffic situation. Thirdly, we are surprised to find that GPT-4V can comprehend dooring and red-light running cases in a zero-shot manner, and make a reasonable analysis of the causes and responsibility of these accidents. Cause reasoning and responsibility reasoning take the successful identification of the accident type as the prerequisite, and require the accurate recognition of the details, as well as cross-modal high-level reasoning capabilities. This suggests that GPT-4V has been able to acquire the cross-modal semantic associations of these cases, showing technical possibilities for other more complex traffic events.

\begin{table}[h]
\centering
\resizebox{\textwidth}{!}{%
\begin{tabular}{c|>{\centering\arraybackslash}p{1.8cm}|>{\centering\arraybackslash}p{1.8cm}|>{\centering\arraybackslash}p{1.8cm}|>{\centering\arraybackslash}p{1.8cm}|>{\centering\arraybackslash}p{1.8cm}|>{\centering\arraybackslash}p{1.8cm}|>{\centering\arraybackslash}p{1.8cm}}    
\toprule
\textbf{Traffic Event} & \textbf{Accident / Event Recognition}  & \textbf{Event Type Identification} & \textbf{Severity Analysis} & \textbf{Emergent Decision Making} & \textbf{Involved Num. Participants} & \textbf{Cause Attribution} & \textbf{Responsibility Reasoning} \\\midrule
Dooring \ref{Dooring} & \checkmark & \checkmark & \checkmark & \checkmark & \checkmark & \checkmark & \checkmark \\\midrule
Run the Red Light at Night \ref{Run the Red Light at Night} & \checkmark & \checkmark & \checkmark & \checkmark & \checkmark & \checkmark & \checkmark \\\midrule
Motorcycle Car Collision \ref{Motorcycle Car Collision} & \checkmark & \checkmark & \checkmark & \checkmark & \checkmark & - & - \\\midrule
Rollover \ref{Rollover}& \checkmark & \checkmark & \checkmark & \checkmark & \checkmark & \XSolidBrush & - \\\midrule
Fires and Explosion \ref{Fires and Explosion} & \checkmark & \checkmark & \checkmark & \checkmark & \XSolidBrush & \XSolidBrush & - \\\midrule
Vehicle Collision \ref{Vehicle Collision} & \XSolidBrush & - & - & - & - & - & - \\\midrule
Vehicle Collision (Highlighted) \ref{Vehicle Collision} & \checkmark & \checkmark & \checkmark & \checkmark & \XSolidBrush & \XSolidBrush & \XSolidBrush \\\midrule
Jaywalking \ref{Jaywalking} & \XSolidBrush & \checkmark & - & - & \XSolidBrush & - & - \\\midrule
Vehicle Smoke \ref{Vehicle Smoke} & \XSolidBrush & \XSolidBrush & - & - & - & - & - \\\midrule
Vehicle Smoke (Highlighted) \ref{Vehicle Smoke} & \XSolidBrush & \XSolidBrush & - & - & - & - & - \\\midrule
Multiple Vehicle Collision \ref{Multiple Vehicle Collision} & \checkmark & \XSolidBrush & - & - & - & - & - \\\bottomrule
\end{tabular}
}
\caption{Evaluation Summary of all Traffic Events.}
\label{evaluate_table}
\end{table}

We summarize and analyze the failed cases and answers, and divide the failure reasons into two main types, i.e. performance limitations and inherent restrictions. Performance limitations can be alleviated with either the scaling up of the dataset or the improvement of training and deployment techniques. Whereas, the inherent restrictions cannot be resolved with the current GPT-4V model, and require the support of extra modalities. \\

\textbf{Performance Limitation} consists of the following aspects. Firstly, the spatial reasoning ability of GPT-4V is weak. As mentioned in \ref{Run the Red Light at Night}, \ref{Rollover}, \ref{Motorcycle Car Collision}, the poor spatial reasoning leads to mistakes when it comes to describing the location of the vehicle, the vehicle's orientation, and the relative positions to other traffic participants. This greatly affects the accuracy of the generated accident reports that involve spatial descriptions. Secondly, GPT-4V is still not accurate enough in recognizing the details of objects of interest, including the color of the vehicles, whether the vehicle tows a trailer, and whether the motorcyclist wears a helmet. These details play an important role in the reasoning of the accident causes, in generating precise descriptions of vehicles involved in the traffic events, and in the analysis of accident severity. Thirdly, GPT-4V suffers from poor cross-image associations of multiple objects, especially when there is dense traffic in the images, or when cars with similar appearance come close to each other. This leads to the failure to provide a precise description of the trajectories and behaviors of target vehicles. \\

\textbf{Inherent Restrictions} refers that only utilizing the input modalities supported by GPT-4V, i.e. texts and images, can not provide the minimum required information for successful recognition and reasoning. For example, it is difficult to identify the explosion without acoustic information, and it is also challenging to recognize a direct collision between vehicles only through images, especially when the vehicle's direction is parallel to the camera's viewpoint. All of this points to the fact that the model could tackle these problems only with the support of additional modalities and additional information, including acoustic, videos, and 3D spatial information.

\newpage
\printbibliography


\end{document}